\documentclass{svproc}
\usepackage{url}
\usepackage{graphicx}
\usepackage{microtype}
\usepackage{multirow}
\usepackage{hhline}
\usepackage{tablefootnote}
\usepackage[hidelinks]{hyperref}

\begin{document}
\mainmatter              
\title{Evaluation of Representation Models for Text Classification with AutoML Tools}
\titlerunning{Representations for Text Classification with AutoML}  

\author{Sebastian Brändle\inst{1} \and Marc Hanussek\inst{1} \and Matthias Blohm\inst{1} \and Maximilien Kintz\inst{2}}

\authorrunning{Sebastian Brändle et al.} 

\tocauthor{Sebastian Brändle, Marc Hanussek, Matthias Blohm, Maximilien Kintz}
\institute{University of Stuttgart IAT, Institute of Human Factors and Technology Management, Stuttgart, Germany,\\
\email{sebastianbraendle@gmx.de},\\ WWW home page:
\texttt{https://www.iat.uni-stuttgart.de/en/}
\and
Fraunhofer IAO, Fraunhofer Institute for Industrial Engineering IAO,\\
Stuttgart, Germany}

\maketitle              

\begin{abstract}
Automated Machine Learning (AutoML) has gained increasing success on tabular data in recent years. However, processing unstructured data like text is a challenge and not widely supported by open-source AutoML tools. This work compares three manually created text representations and text embeddings automatically created by AutoML tools. Our benchmark includes four popular open-source AutoML tools and eight datasets for text classification purposes. The results show that straightforward text representations perform better than AutoML tools with automatically created text embeddings.
\keywords{AutoML, NLP, Text Classification, Machine Learning, Text Representations, Text Embeddings}
\end{abstract}
\section{Introduction}
\label{sec:Introduction}

The basic idea of Automated Machine Learning is to find a suited machine learning (ML) algorithm, preprocessing of input features, and set of hyperparameters given a dataset and a computational budget. More generally speaking, AutoML seeks to automate the time-consuming and tedious tasks when developing machine learning models. A lot of benchmarking work has been done for measuring performance of today's AutoML tools, especially on tabular datasets~\cite{He2021AutoMLAS}~\cite{confictaiTruongWGHBF19}~\cite{zoeller2021benchmark}. On the other hand, usage of AutoML for tasks of natural language processing (NLP), e.g. text classification or named entity recognition, is still in its very early stages. Nonetheless, we experienced growing interest and need for solutions in the area of automated text categorization or analysis from many of our customers and enterprises in general, especially for supporting processing and routing of daily incoming textual documents like letters, emails, invoices etc. However, building suitable NLP machine learning applications for specific use cases and document types is not trivial, since implementation of such solutions usually requires deeper knowledge about appropriate text preprocessing and model building techniques, which could be facilitated by the use of AutoML tools.

While many popular cloud providers like Google, Microsoft, IBM or Amazon already have built-in AutoML functions for NLP, many open-source AutoML libraries have no support for processing raw text input yet. For still being able to use those tools for aims like text classification, training data in form of text input has to be converted to tabular data by hand. Common representation techniques for word or sentence embeddings (embeddings, in short) like Word2Vec~\cite{mikolov2013efficient} or textual representations created from BERT~\cite{BERT} models are obvious choices for experiments in this area.

In this work we evaluate the performance of various AutoML approaches on the task of text classification for eight common English textual datasets. We compare performances achieved by different text embeddings and AutoML tools. We are not aware of existing benchmarks that investigate the suitability of different text embeddings for text classification by AutoML tools, thus we contribute to the current state of knowledge in this field.

The paper is structured as follows: In \autoref{sec:Related Work} we list related work in the field of AutoML applied for NLP tasks. In \autoref{sec:Methodology}, the methodology and settings for our experiments are described. Discussion and analysis of our results are given in \autoref{sec:Results}, followed by a conclusion and description of relevant future work in \autoref{sec:Conclusion}.

\section{Related Work}
\label{sec:Related Work}

Popular providers of AI solutions like Google, from where the commonly used BERT-representations~\cite{BERT} originally emerged, provide AutoML-based services for NLP tasks like text classification, named entity recognition and sentiment detection within their platforms\footnote{https://cloud.google.com/natural-language/automl/}. Similar products of other big companies like Amazon Comprehend~\cite{Comprehend}, that came up within the last years and that also provide AutoML services for NLP tasks, underline the growing relevance of this topic. However, though those tools achieve good results for many cases, using cloud services on sensitive in-house data like customer correspondence letters is usually not possible without issues, which often make open-source or on premise deployable solutions the better choice.  On the other hand, within the domain of open-source technology, support for automatic processing of raw text input is still in its very early stages when writing this paper, and many popular tools like Auto-Sklearn~\cite{Auto-Sklearn_Paper} still lack such capabilities. 

Nonetheless, individual works like indeed focus on automation of machine learning tasks for NLP by proposing general methods for raw text pre processing and vector transformation~\cite{madrid2019meta-learning}. 
Further experiments investigated usage of transfer learning for automated text categorization with promising results~\cite{NEURIPS2018_bdb3c278}. Other approaches use language embeddings created from a dataset's natural language meta data (like data set description texts) for improving performance in AutoML trainings~\cite{journals/corr/abs-1910-03698}.

Additionally, benchmarks have been done that compare several AutoML approaches and their performance on different problems, also including NLP tasks~\cite{He2021AutoMLAS}. For instance, in one of our previous works we have performed an open-source AutoML benchmark for text classification~\cite{LaveragingAutoML}. The focus was on comparing the classification results of four different AutoML tools both among each other and with the results of machine learning experts. The text representations for the AutoML programs were generated using a transformer tool based on the BERT model. In contrast, the present paper focuses on the comparison of the classification performance of manually generated text representations and internally generated text representations (that is, embeddings that are automatically calculated by AutoML tools as part of the preprocessing). In detail, coming from a practical point of view and unlike most other related work in this field, our main goal here was to find well performing text representations that are easy to compute and which can be handled well by nowadays open-source AutomML tools without any further configuration or adaptation.   

\section{Methodology of the Benchmark}
\label{sec:Methodology}
\begin{figure}[hbt!]
   \centering
   \includegraphics[width=\textwidth]{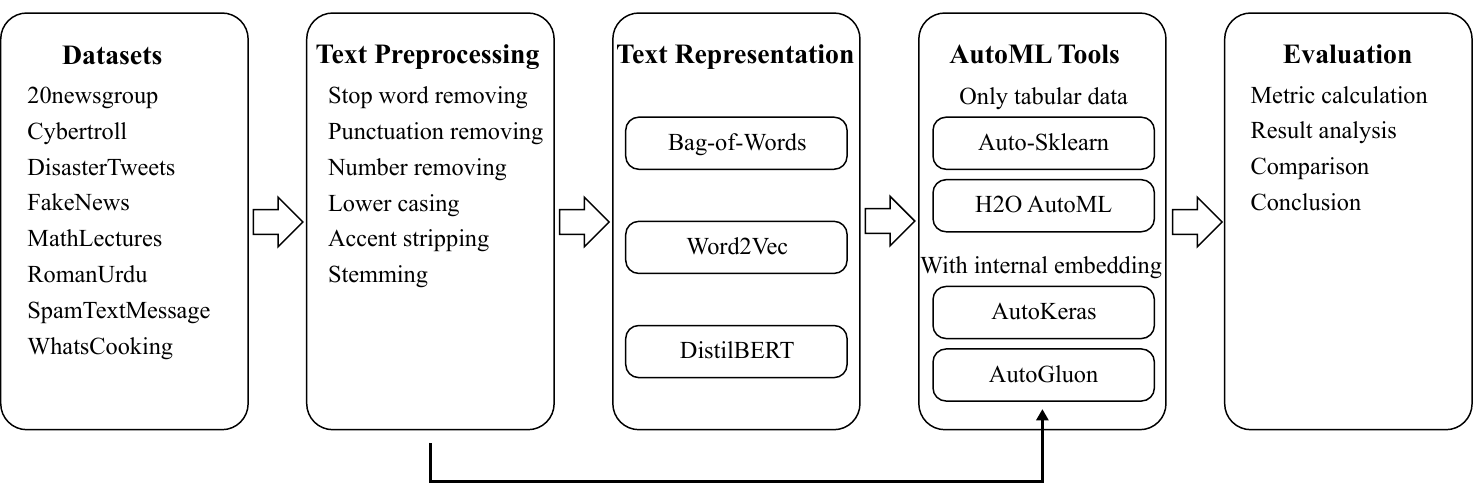}
   \caption{Overview of the approach of the present benchmark}
   \label{fig:ApproachOverview}
\end{figure}

The following chapter describes the underlying methodology and assumptions of the present benchmark. As can be seen in \autoref{fig:ApproachOverview}, the used approach can be divided into five steps. The first step deals with the selection of datasets for classification tasks. Then standardized text data preprocessings such as removal of stop words or lower casing is performed. In a third step, each dataset is processed with three different text representation models (see \autoref{subsec:Data Representation} for their description). This provides a tabular-shaped representation of the datasets and enables their use by AutoML tools. The fourth step contains the classification with four different AutoML tools. This includes the processing of the represented datasets from the previous step as well as the processing of raw text (only possible with two AutoML tools). In the final step the classification results of each tool with the corresponding text representation model is evaluated.  

\subsection{Assumptions}
\label{subsec:Assumptions}
The present benchmark follows a practical point of view and considers AutoML as a technical solution which enables non-experts to make use of ML. This includes the assumption that the user is a domain expert with minor experience in ML and understands AutoML as a tool to solve a business problem at hand. Therefore, the AutoML tools are consequently treated as "black boxes". When the tools work without interruption or error, all settings have been left unaltered. 

\subsection{Datasets and Data Preparation}
\label{subsec:Datasets}
As shown in \autoref{tab:DatasetsOverview}, eight publicly available text datasets for classification purposes are considered. This includes a variety of tasks like sentiment analysis, fake news, and spam message detection. Five out of eight datasets account for binary classification. The \emph{20newsgroup}, \emph{MathLectures} and \emph{WhatsCooking} datasets stand for multiclass classification tasks with 20, 11 and 20 classes, respectively. The benchmark focuses on small and medium sized datasets which cover a span between 804 (\emph{FakeNews}) and 39774 (\emph{WhatsCooking}) samples. For convenience, the best-known results are given.

To prepare the datasets in a unified way, all text documents are preprocessed with a standardized script that applies basic NLP techniques like stop word, punctuation and number removing as well as lower casing, accent stripping and stemming. All samples that no longer contain text after preprocessing are removed from the dataset. The vocabulary of the remaining datasets covers a wide range from only 2575 distinct words for the \emph{WhatsCooking} dataset to 152671 words for the \emph{20newsgroup} dataset. Also the average document length varies widely between seven words (\emph{Cybertroll}) and 2122 words (\emph{MathLectures}). 

\subsection{Text Data Representation}
\label{subsec:Data Representation}
To cover a wide variety of text representation methods, one model of each popular model group was implemented. The considered model groups are n-gram models, static embeddings, and dynamic embeddings. Due to the fact that n-gram models as well as static word embeddings can achieve good results for classification tasks~\cite{ConvolutionalNeuralNetworksForSentenceClassification}~\cite{MachineLearningForText}~\cite{ComparisonOfCountBasedAndPredictionBasedVectorizers} it is an additional point to take them into account. For this reason, beside the most recent context dependent embeddings, also two context independent text representations are part of the present benchmark. 

All three text representation models were implemented in a straightforward way. Neither fine-tuning of the models nor adjustment according to the datasets (or their domains) have been made. It is known that the model quality can rely significantly on fine-tuning and case-specific adaptations, but, according to the AutoML spirit, we abstain from assuming such deeper knowledge on the part of the target group. The three different representations are called \emph{external} generated text representations in the following.

\subsubsection{N-gram Model (Bag-of-Words)}
The n-gram text representation is realized by a unigram or also called Bag-of-Words model and therefore, only one word at a time is considered. To build the corresponding vocabulary, the entire corpus of each dataset is used. The vectorization is performed by a simple count vectorizer and neither term frequency nor normalization is applied.


\subsubsection{Static Embedding (Word2Vec)} 
The implementation of static or context independent embeddings is achieved using pretrained word vectors. These word vectors are obtained from a publicly available Word2Vec model\footnote{GoogleNews-vectors-negative300, \url{https://code.google.com/archive/p/word2vec/}}~\cite{mikolov2013efficient}, trained on the Google News text corpus. The method to calculate the embeddings is a data driven approach and described in~\cite{mikolov2013distributed}. Each word is represented by a 300-dimensional vector.

To represent the considered datasets with the pretrained word embeddings, two data processing steps were performed. The first step contains the comparison of occurring words. Every word token of the documents of each dataset is compared with the available pretrained word embeddings obtained from the model. If a word or token cannot be represented due to a missing pretrained embedding, it is removed from the text corpus. Only those words that can be represented by a pretrained embedding from the Word2Vec model are retained. In the second step the three-dimensional document embedding is reduced to a two dimensional tabular representation. Therefore, the average of all word vector embeddings, which are representing a single document, is calculated. The averaged word embeddings represent the entire document with one 300-dimensional vector. This allows a tabular shaped representation of the individual datasets, which can be used as input data for AutoML tools.

Hereafter, the described approach using averaged static word vectors is referred as ''W2V-Average'' embedding. 

\subsubsection{Dynamic Embedding (DistilBERT)}
For the creation of dynamic or context dependent word embeddings, with BERT~\cite{BERT} one of the most popular language models is used. To reduce the required computational resources a distilled version of the BERT model is applied. This so called DistilBERT model is 40\,\% smaller and 60\,\% faster as the original BERT model and achieves still about 97\,\% of its initial language understanding~\cite{distilBert}. 

The context dependent embeddings of the dataset documents are generated by a pretrained DistilBERT model\footnote{'distilbert-base-uncased' model, \url{https://huggingface.co/transformers/model_doc/distilbert.html\#distilbertmodel}}. In a first step a tokenizer function is used to prepare the documents for the DistilBERT model. The maximum number of tokens per document is limited to 512. Therefore, longer documents are truncated, and shorter documents are extended with zero padding. The documents, prepared with the tokenizer function, are the input of the DistilBERT model. The output of the DistilBERT model provides, among others, a special classification token, which represents the entire document~\cite{BERT}. This [CLS]-token represents each document with a 768-dimensional embedding vector. 



\subsubsection{Internal Embeddings}
Two of the four considered AutoML tools provide the possibility to process raw text automatically\footnote{This is why the last column in \autoref{tab:Results} lacks every other value.}. AutoKeras as well as AutoGluon are able to handle raw text, create embeddings and perform the classification task without human interaction. 

The TextPredictor class of AutoGluon, which is capable of raw text processing, only uses transformer neural network models. Pretrained language models like BERT, ALBERT and ELECTRA are fit to the input data via transfer learning\footnote{AutoGluon Documentation, \url{https://auto.gluon.ai/stable/tutorials/text_prediction/beginner.html}}. 


\subsection{AutoML Tools}
\label{subsec:AutoML Tools}
Four open-source AutoML tools are considered for the present benchmark. Auto-Sklearn\footnote{Auto-Sklearn Version: 0.11.1}~\cite{Auto-Sklearn_Paper} as well as H2O AutoML\footnote{H2O AutoML Version: 3.32.0.2}~\cite{H2OAutoML20} can process tabular data for classification tasks. Therefore, with both tools it is only possible to perform a text classification if the text data is prepared beforehand (as described in \autoref{subsec:Data Representation}). In contrast, AutoGluon\footnote{AutoGluon Version: 0.0.15}~\cite{AutoGluon_Paper} and AutoKeras\footnote{AutoKeras Version: 1.0.12}~\cite{AutoKeras} are both able to perform classification tasks with tabular and raw text data. 

To compare the different text representation methods, every AutoML tool is used to perform a classification task with all three external created text representations. This includes the Bag-of-Words model, the W2V-Average embedding and the DistilBERT embedding for each dataset. In addition, the raw text processing ability of AutoGluon and AutoKeras is used to perform classification tasks with automatically and internally produced text embeddings. 

The time limit for the calculation is set to 60 minutes per fold for each AutoML tool. Only AutoKeras requires manual adjustment of the settings due to the missing possibility of specifying a time limit. Clearly, this can affect the comparability of the tools.



\subsection{Cross-Validation and Metrics}
\label{subsec:cvandmetrics}
The results of all eight datasets are obtained by a 5-fold cross validation. This includes random shuffling and stratification for each train-test-split. The primary optimization and evaluation metric is accuracy. The final evaluation scores are calculated as the average of the five folds.


\subsection{Hardware}
\label{subsec:Hardware}

The workstation, on which the benchmark calculations are performed, is a locally hosted dedicated server. The server is equipped with two Intel Xeon Silver 4114 CPUs @2.20Ghz (yielding 20 cores in total), four 64GB DIMM DDR4 Synchronous 2666MHz memory modules and two NVIDIA GeForce GTX 1080 Ti (yielding more than 22GB VRAM in total).

\section{Results and Analysis}
\label{sec:Results}
\begin{figure}[!htb]
   \centering
   \includegraphics[width=\textwidth]{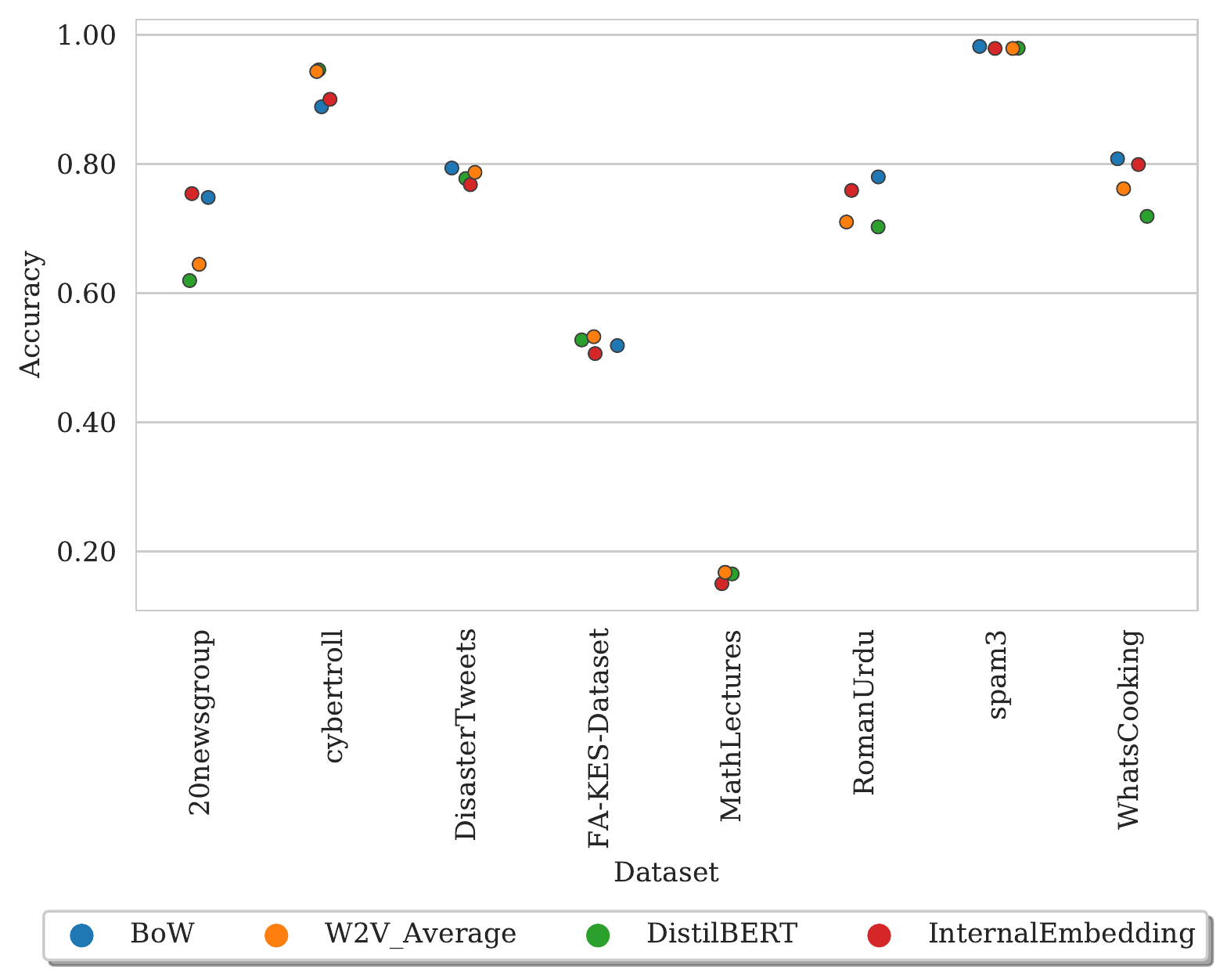}
   \caption{Average classification results of the best text representation for each dataset}
   \label{fig:Stripplot_Datasets_Best}
\end{figure}

The results of the benchmark calculation are shown in \autoref{fig:Stripplot_Datasets_Best}. For reasons of clarity, AutoML tools were excluded from the visualization. On the y-axis the best result (i.e. AutoML tool) of each text representation method is displayed. 

\autoref{tab:Best_Embedding} contains the corresponding mean accuracy values of the data points of \autoref{fig:Stripplot_Datasets_Best}. Through the standardized text preprocessing (\autoref{subsec:Datasets}) and application of text representation models (\autoref{subsec:Data Representation}) all AutoML Tools have the same initial situation. Because of the constant boundary conditions, it is possible to observe differences in the text representation ability of each representation model.

\begin{table}[!htb]
\centering
\caption{Benchmark accuracy results of the best performing AutoML tool of each text representation model and dataset}
\begin{tabular}{|l|r|r|r|r|}
\hline
 \multirow{2}{*}{\textbf{Datasets}} & \multicolumn{4}{c|}{\textbf{Accuracy [\%]}} \\ 
 \cline{2-5}
 & Bag-of-Words & W2V Average & DistilBERT & Internal Embedding \\ \hline
20Newsgroups   & 74.79 & 64.45 & 61.93 & \textbf{75.39} \\ \hline
Cybertroll    & 88.83 & 94.29 & \textbf{94.54} & 89.99 \\ \hline
DisasterTweets & \textbf{79.35} & 78.68 & 77.72  & 76.77 \\ \hline
FakeNews     & 51.86  & \textbf{53.23}  & 52.74 & 50.63  \\ \hline
MathLectures   & 16.74  & \textbf{16.74} & 16.51 & 15.00 \\ \hline
RomanUrdu     & \textbf{77.98} & 70.99 & 70.24  & 75.88 \\ \hline
SpamTextMessage & \textbf{98.17} & 97.87 & 97.90 & 97.87  \\ \hline
WhatsCooking   & \textbf{80.79} & 76.15 & 71.87  & 79.89  \\ \hline
\end{tabular}
\label{tab:Best_Embedding}
\end{table} 

The results obtained by the Bag-of-Words model are best in four out of eight datasets. This means that the simplest approach outperforms the other representation models in 50\,\% of the cases. The second most frequent best results are calculated in combination with the W2V-Average embedding: In 25\,\% of the considered datasets, the classification accuracy is higher than the results of the other text representation models. Only for one dataset each the DistilBERT embedding and the internally generated embeddings manage to achieve a best result. 

Regarding performance of AutoML tools the outcome of the benchmark is shown in \autoref{tab:Best_Tool}. Due to the fact that each dataset is represented by three different, externally generated and one internally generated text representation, every dataset has four best results (one for each representation). \autoref{tab:Best_Tool} shows how many prediction models with the best performance were calculated by each AutoML tool. In total, we report 32 best results for the considered eight datasets.

The results show that AutoGluon calculates the most prediction models outperforming other tools' models: With 56.25\,\% AutoGluon delivers three times as often a best model than Auto-Sklearn or AutoKeras. H2O AutoML achieves two out of 32 best prediction models.

\begin{table}[!htb]
\centering
\caption{Amount of best performing classification models per AutoML tool and dataset}
\begin{tabular}{|l|r|r|r|r|}
\hline
\textbf{Datasets} & \textbf{Auto-Sklearn} & \textbf{AutoKeras} & \textbf{H2O AutoML} & \textbf{AutoGluon} \\ \hline
20Newsgroups   & 3 & 1 & 0 & 0 \\ \hline
Cybertroll    & 0 & 1 & 0 & 3 \\ \hline
DisasterTweets & 0 & 0 & 0 & 4 \\ \hline
FakeNews     & 1 & 1 & 2 & 0 \\ \hline
MathLectures   & 1 & 0 & 0 & 3 \\ \hline
RomanUrdu     & 1 & 1 & 0 & 0 \\ \hline
SpamTextMessage & 0 & 1 & 0 & 3 \\ \hline
WhatsCooking   & 0 & 1 & 0 & 3 \\ \hhline{|=|=|=|=|=|}
Sum     & 6 & 6 & 2 & 18 \\ \hline
Percentage    & 18.75\,\% & 18.75\,\% & 6.25\,\% & 56.25\,\% \\ \hline
\end{tabular}
\label{tab:Best_Tool}
\end{table}

Summing up, it is possible to achieve competitive results with simple text representation approaches. In 75\,\% of the cases context independent text representations (Bag-of-Words and the W2V-Average embedding) provide the best classification accuracy. In contrast, by using context sensitive models (DistilBERT and internal embeddings) only two best results can be calculated. Especially AutoML tools which can create embeddings from raw text are not able to compete with manually generated text representations. From the AutoML tool point of view, AutoGluon outperforms the competitors by far.

\section{Conclusion and Outlook}
\label{sec:Conclusion}
The overall goal of the present work is to investigate AutoML performance on text classification tasks. This research considers both the influence of text representation models on the classification quality of the AutoML tools as well as the performance of the AutoML tools for their part. However, the focus is on text representation models. The evaluation of the benchmark shows that simple text representation methods like Bag-of-Words achieve competitive results while fully automated text embeddings are not yet able to beat manually created text representations. 

Regarding the limitations of this work, there are open questions that are worth exploring in the future. An important extension of the present work is to increase both the amount of considered datasets as well as datasets' sample sizes. Such an extension could help confirm the present observations and give additional insights, for example under which circumstances advanced embeddings play out strength. Another topic we did not cover in the present work is a dataset specific adaption of text preprocessing and fine-tuning of text representation models. Furthermore, there are lots of open-source AutoML tools other than the considered four, which should be taken into account. 

We were surprised by the fact that simple text embeddings such as Bag-of-Words perform best and are able to outperform more recent text embeddings. This shows that text classification with AutoML, at least in the context of our benchmark, is different from standard text classification in terms of data preprocessing. In our opinion, AutoML tools with native raw text support are of high interest, as they are very user-friendly and may provide even better results in the future.

\bibliographystyle{bibtex/spmpsci_unsrt}
\bibliography{paper}

\newpage
\section*{Appendix}

\begin{table}
\centering
\caption{Datasets overview}
\begin{tabular}{|l|p{6cm}|r|r|r|}
\hline
\textbf{Dataset name} & \textbf{Source and/or Reference} & \textbf{Classes} & \textbf{Samples} & \textbf{\parbox{2cm}{Best-known result \\(Accuracy [\%])}}\\
\hline
20Newsgroups & \url{https://scikit-learn.org/0.19/datasets/twenty_newsgroups.html} & 20 & 18333 & 86.8~\cite{yamada-shindo-2019-neural} \\ \hline
Cybertroll & \url{https://zenodo.org/record/3665663} & 2 & 20001 & 94.4 \tablefootnote{\url{https://www.kaggle.com/kevinlwebb/cybertrolls-exploration-and-ml}} \\ \hline
DisasterTweets & \url{https://www.kaggle.com/c/nlp-getting-started/data} & 2 & 7613 & 100 \tablefootnote{\url{https://www.kaggle.com/c/nlp-getting-started/leaderboard}} \\ \hline
FakeNews & \url{https://www.kaggle.com/mohamadalhasan/a-fake-news-dataset-around-the-syrian-war} & 2 & 804 & 51.9 \tablefootnote{\url{https://www.kaggle.com/mohamadalhasan/fake-news-around-syrian-war}}  \\ \hline
MathLectures & \url{https://www.kaggle.com/extralime/math-lectures} & 11 & 860 & 16.9 \tablefootnote{\url{https://www.kaggle.com/sanskar27jain/kernelb49fc09b70}}  \\ \hline
RomanUrdu & \cite{RomanUrdu} & 2 & 10185 & 87~\cite{RomanUrdu} \\ \hline
SpamTextMessage & \url{https://www.kaggle.com/team-ai/spam-text-message-classification} \cite{SpamFiltering} & 2 & 5555 & 96.6 \tablefootnote{\url{https://www.kaggle.com/vennaa/notebook-spam-text-message-classification-with-r}} \\ \hline
WhatsCooking & \url{https://www.kaggle.com/c/whats-cooking/overview} & 20 & 39774 & 83.2 \tablefootnote{\url{https://www.kaggle.com/c/whats-cooking/leaderboard}}\\ \hline
\end{tabular}
\label{tab:DatasetsOverview}
\end{table}

\begin{table}
\centering
\caption{Average accuracy results of the benchmark}
\begin{tabular}{|l|l|r|r|r|r|}
\hline
 \multirow{2}{*}{\textbf{Datasets}} & \multirow{2}{*}{\textbf{AutoML Tool}} & \multicolumn{4}{c|}{\textbf{Accuracy [\%]}} \\ 
 \cline{3-6}
& & Bag-of-Words & W2V Average & DistilBERT & Internal Embedding \\ \hline
\multirow{4}{*}{20Newsgroups}  & Auto-Sklearn & 74.79 & 64.45 & 61.93 & -       \\ \cline{2-6}
                     & AutoKeras   & -      & 61.12 & 53.52 & 75.39 \\ \cline{2-6}
                     & H2O AutoML  & 50.12 & 62.64 & 61.07 & -       \\ \cline{2-6}
                     & AutoGluon   & -      & 64.01 & 60.89 & 43.83 \\ \hhline{|=|=|=|=|=|=|}
\multirow{4}{*}{Cybertroll}   & Auto-Sklearn & 80.64 & 89.56 & 89.97 & -       \\ \cline{2-6}
                     & AutoKeras   & -      & 89.57 & 83.67 & 89.99 \\ \cline{2-6}
                     & H2O AutoML  & 83.05 & 91.40 & 92.12 & -       \\ \cline{2-6}
                     & AutoGluon   & 88.83 & 94.29 & 94.54 & 72.02 \\ \hhline{|=|=|=|=|=|=|}
\multirow{4}{*}{DisasterTweets} & Auto-Sklearn & 79.11 & 77.92 & 77.71 & -       \\ \cline{2-6}
                     & AutoKeras   & -      & 77.25 & 73.41 & 75.72 \\ \cline{2-6}
                     & H2O AutoML  & 76.01 & 74.68 & 73.69 & -       \\ \cline{2-6}
                     & AutoGluon   & 79.35 & 78.68 & 77.72 & 76.77 \\ \hhline{|=|=|=|=|=|=|}
\multirow{4}{*}{FakeNews}     & Auto-Sklearn & 51.86 & 50.75 & 51.74 & -       \\ \cline{2-6}
                     & AutoKeras   & -      & 51.36 & 50.61 & 50.63 \\ \cline{2-6}
                     & H2O AutoML  & 51.37 & 53.23 & 52.74 & -       \\ \cline{2-6}
                     & AutoGluon   & 50.63 & 49.13 & 52.61 & 49.76 \\ \hhline{|=|=|=|=|=|=|}
\multirow{4}{*}{MathLectures}  & Auto-Sklearn & 15.81 & 15.47 & 16.51 & -       \\ \cline{2-6}
                     & AutoKeras   & -      & 13.37 & 11.63 & 9.19 \\ \cline{2-6}
                     & H2O AutoML  & 16.26 & 11.05 & 11.28 & -       \\ \cline{2-6}
                     & AutoGluon   & 16.74 & 16.74 & 16.51 & 15.00 \\ \hhline{|=|=|=|=|=|=|}
\multirow{4}{*}{RomanUrdu}    & Auto-Sklearn & 77.98 & 69.52 & 69.93 & -       \\ \cline{2-6}
                     & AutoKeras   & -      & 66.53 & 67.64 & 75.88 \\ \cline{2-6}
                     & H2O AutoML  & 75.17 & 66.85 & 65.88 & -       \\ \cline{2-6}
                     & AutoGluon   & 77.65 & 70.99 & 70.24 & 71.14 \\ \hhline{|=|=|=|=|=|=|}
\multirow{4}{*}{SpamTextMessage}& Auto-Sklearn & 97.43 & 96.55 & 97.83 & -       \\ \cline{2-6}
                     & AutoKeras   & -      & 97.18 & 97.51 & 97.87 \\ \cline{2-6}
                     & H2O AutoML  & 97.62 & 97.36 & 96.47 & -       \\ \cline{2-6}
                     & AutoGluon   & 98.17 & 97.87 & 97.90 & 97.18 \\ \hhline{|=|=|=|=|=|=|}
\multirow{4}{*}{WhatsCooking}  & Auto-Sklearn & 75.90 & 74.61 & 71.38 & -       \\ \cline{2-6}
                     & AutoKeras   & -      & 73.83 & 65.99 & 79.89 \\ \cline{2-6}
                     & H2O AutoML  & 78.50 & 74.51 & 70.02 & -       \\ \cline{2-6}
                     & AutoGluon   & 80.79 & 76.15 & 71.87 & 65.88 \\ \hline
\end{tabular}
\label{tab:Results}
\end{table}
\end{document}